\documentclass[runningheads]{llncs}

\usepackage{graphicx}
\usepackage[utf8]{inputenc}
\usepackage[T1]{fontenc}
\usepackage{amssymb,amsmath,array}
\usepackage{wrapfig}
\usepackage{hyperref}
\usepackage{url}
\usepackage{booktabs}
\usepackage{amsfonts}
\usepackage{nicefrac}
\usepackage{microtype}
\usepackage{xcolor}
\usepackage{makecell}
\usepackage{placeins}
\usepackage{float}

\begin{document}

\title{High-Fidelity Mural Restoration via a Unified Hybrid Mask-Aware Transformer}

% If the paper title is too long for the running head, you can set an abbreviated paper title here
\titlerunning{Hybrid Mask-Aware Transformer for Mural Restoration}

% \author{Anonymous submission}
% \institute{}
\author{Jincheng Jiang\inst{1} \and
Qianhao Han\inst{2} \and
Chi Zhang\inst{1} \and
Riddhi Shyambhushan Mandal\inst{1} \and
Zheng Zheng\inst{1}}
\authorrunning{J. Jiang et al.}
\institute{Northeastern University (Toronto Campus), Toronto, Canada \\
\email{\{jiang.jinc, zhang.chi27, mandal.rid, zh.zheng\}@northeastern.edu} \and
The Bishop Strachan School, Toronto, Canada\\
\email{qianhaoh27@bss.on.ca}}

\maketitle              % typeset the header of the contribution

\begin{abstract}
Ancient murals are valuable cultural artifacts, but many have suffered severe degradation due to environmental exposure, material aging, and human activity. Restoring these artworks is challenging because it requires both reconstructing large missing structures and preserving authentic, undamaged regions. We present the Hybrid Mask-Aware Transformer (HMAT), a unified framework for high-fidelity mural restoration that addresses both structural completion and authentic-region preservation. HMAT integrates Mask-Aware Dynamic Filtering for robust local texture modeling with a Transformer bottleneck for long-range structural inference, enabling recovery of continuous line patterns and coherent mural structures under irregular damage. To handle diverse degradation morphologies, we introduce a mask-conditional style fusion module that adapts the generative process according to the shape and extent of missing regions. We also propose a fidelity-oriented training objective that combines hole-normalized reconstruction, discriminator feature matching, and high-receptive-field perceptual supervision to improve damaged-region fidelity, texture consistency, and boundary quality. In addition, we analyze a Teacher-Forcing Decoder with hard-gated skip connections as a feature-space boundary-conditioning strategy. Experiments show that HMAT matches or outperforms representative convolutional, transformer-based, and edge-guided inpainting baselines, with especially strong gains in perceptual realism and severe-mask settings. Ablation studies further identify the proposed objective, MADF-based mask-aware encoding, and mask-conditioned synthesis as the main contributors to restoration quality. These results demonstrate that HMAT provides an effective and competitive solution for cultural heritage mural restoration.

\end{abstract}

\section{Introduction}\label{intro}
Ancient murals often suffer severe deterioration due to environmental exposure, biological decay, and human activity~\cite{meijun15}, typically manifesting as cracks, paint loss, mold, or a combination of these, which poses substantial challenges for restoration. The task is further complicated by the need to preserve not only visual elements such as texture and color, but also the historical and aesthetic significance embedded in the artwork. Traditional image restoration approaches are often inadequate for addressing the complex and overlapping degradation patterns typically present in murals.

Although deep learning has significantly advanced image restoration~\cite{zhang2023image,xiang2023deep,quan2024deep}, its application to mural restoration remains challenging. First, murals often exhibit both fine-grained degradation (e.g., cracks and scratches) and large-scale structural loss (e.g., extensive peeling)~\cite{meijun15}. Second, the visual characteristics of murals differ substantially from natural images, making it difficult for models trained on standard datasets to generalize effectively~\cite{li2022computing}. Third, cultural heritage restoration requires preservation of undamaged regions, while existing methods may still introduce artifacts or color shifts near authentic content.

To address the above challenges, we propose the Hybrid Mask-Aware Transformer (HMAT), a unified framework for automated mural restoration. Rather than relying on a single architectural change, HMAT combines three complementary components: a hybrid mask-aware generator, a mask-conditioned synthesis mechanism, and a fidelity-oriented training objective. Architecturally, HMAT integrates MADF~\cite{Zhu_2021} for mask-aware local texture modeling with a Transformer~\cite{li2022matmaskawaretransformerlarge} bottleneck for long-range structural inference, while a mask-conditional style fusion module adapts synthesis to the morphology of degradation, from thin cracks to extensive peeling. Authentic, undamaged pixels are retained exactly through output-level compositing. Importantly, our ablation study shows that the training objective is a decisive factor for high-fidelity mural restoration: coupling a hole-normalized reconstruction loss with discriminator feature matching and a high-receptive-field perceptual loss substantially improves fidelity over the conventional adversarial-plus-VGG objective used by prior inpainting models. The main contributions are summarized as follows:
\begin{itemize}
    \item We propose HMAT, a unified framework for mural restoration that integrates MADF-based local texture modeling, Transformer-based global structural reasoning, mask-conditional style fusion, and fidelity-oriented optimization within a single restoration pipeline.
    \item We design a fidelity-oriented training objective for mural restoration, combining hole-normalized reconstruction, discriminator feature matching, and a high-receptive-field perceptual loss, and show through ablation that it contributes the largest performance gain among the tested components.
    \item We evaluate HMAT on the DHMural and Nine-Colored Deer datasets under varying degradation levels, where it matches or surpasses strong inpainting baselines in both pixel fidelity and perceptual realism while exactly preserving authentic regions.
\end{itemize}

\section{Related Work} \label{relate}
Traditional inpainting algorithms primarily rely on diffusion-based~\cite{bertalmio00} or patch-based strategies~\cite{barnes09}. Diffusion methods propagate local information using partial differential equations but often produce blurred results in complex textures. Conversely, patch-based approaches~\cite{criminisi04} synthesize details by copying similar exemplars from the background. While efficient, these methods lack high-level semantic understanding, making them ineffective for the large-scale, structural damage frequently observed in ancient murals.

Most deep learning-based image restoration methods adopt an encoder-decoder architecture, in which the encoder extracts features from degraded images and the decoder reconstructs the output. Convolution-based methods are effective at modeling local textures, but their limited receptive fields make it difficult to recover large missing structures. To ensure that structural outlines are accurately reconstructed before filling in missing textures, two-stage methods such as EdgeConnect~\cite{nazeri19} were introduced to hallucinate intermediate edge maps. While successful on modern datasets, EdgeConnect relies on external edge detectors that are overly sensitive to image degradation. When applied to ancient murals, they frequently extract noise as false edges and fail to capture true, ambiguous outlines. To avoid relying on explicit edge maps while better handling irregular masks, MADF~\cite{Zhu_2021} utilizes spatially variant convolutions to prevent boundary artifacts and zero-mixing. LaMa~\cite{suvorov2022resolution} further improves large-mask inpainting by using Fourier convolutions with image-wide receptive fields. However, these convolution-oriented methods can still struggle to align distant semantic structures when the missing region crosses long mural contours. Orthogonal to architecture, the training objective strongly affects inpainting quality: discriminator feature-matching and high-receptive-field perceptual losses~\cite{suvorov2022resolution} improve structural coherence and texture realism over standard adversarial and VGG-based supervision. Yet objective design has received less systematic attention than architectural design in mural restoration, where models are often trained with conventional adversarial and perceptual losses.

To overcome the limitations of CNNs, attention-based methods~\cite{yu18,huang2024} have been introduced to model long-range dependencies and global context. In particular, purely attention-based architectures such as MAT~\cite{li2022matmaskawaretransformerlarge} leverage global contextual aggregation to more effectively hallucinate large missing structures. In the cultural heritage domain, ADF~\cite{shao2023} introduces a diffusion-based framework for blind mural restoration on the DHMural benchmark, targeting globally degraded murals rather than mask-conditioned completion of specified regions. However, because attention-heavy methods lack the strong local inductive bias of CNNs, they often struggle to reconstruct sharp, high-frequency textures, particularly under the limited training data typical of cultural heritage datasets. Moreover, most existing methods inherit architectures and objectives designed for natural images and place limited emphasis on valid-region fidelity under the severe, mixed degradations of ancient murals, which limits their applicability to cultural heritage restoration.

% However, due to the lack of strong local inductive bias inherent in CNNs, attention-heavy methods often struggle to reconstruct sharp, high-frequency textures, particularly under the limited training data typical of cultural heritage datasets.

\section{The Proposed Approach} \label{approach}

\subsection{Problem Formulation and Notation}
We formalize mural restoration as a conditional image completion problem with an identity constraint on observed regions~\cite{zhang2023image,xiang2023deep}. Let $x \in \mathbb{R}^{3 \times H \times W}$ represent the ground-truth, undamaged mural image, and let $m \in \{0,1\}^{1 \times H \times W}$ denote the binary visibility mask, where $m(p)=1$ indicates a known, authentic pixel and $m(p)=0$ indicates a missing or damaged pixel. We define the missing region as $\Omega = \{p \mid m(p)=0\}$ and the valid region as $\bar{\Omega} = \{p \mid m(p)=1\}$.

To perform the restoration, HMAT receives a degraded input image, which is formed by preserving only the valid pixels while masking the damaged areas: $x_{obs} = x \odot m$, where $\odot$ denotes element-wise multiplication. Our objective is to train the generator $G$ to produce a restored image $\hat{x}$ that satisfies two key criteria: (1) synthesizing structurally coherent content within the missing regions ($\Omega$), and (2) preserving the observed pixels in the undamaged regions ($\bar{\Omega}$) through final compositing.

To account for the inherent uncertainty of missing regions and enable multi-modal synthesis, we introduce a latent noise vector $z \sim \mathcal{N}(0, I)$, formulating the generative prediction as $G(x_{obs}, m, z)$. Unlike standard image editing, cultural heritage restoration places strong
emphasis on retaining surviving historical paint~\cite{shao2023,li2022computing}. We therefore bypass the generator for known pixels in the final output and produce a composited image:

\begin{equation}
x_{out} = x_{obs} \odot m + G(x_{obs}, m, z) \odot (1 - m)
\label{eq:composite}
\end{equation}

While compositing guarantees pixel-exact fidelity in the valid region $\bar{\Omega}$, it does not constrain the intermediate features used to synthesize content near the mask boundary. This motivates the mask-aware design of HMAT—mask-conditioned encoding and masked contextual aggregation—which limit contamination from zero-filled regions before the final composite is applied.

\subsection{Framework Overview}

As illustrated in Figure~\ref{fig:architecture}, the proposed Hybrid
Mask-Aware Transformer (HMAT) is a unified two-stage generator designed for faithful mural restoration under explicit mask guidance. Given a degraded input $x_{obs} = x \odot m$, a binary mask $m$, and latent noise $z \sim \mathcal{N}(0,I)$, HMAT reconstructs missing regions while retaining observed pixels through output compositing. To handle mixed degradations, from thin cracks to extensive peeling, the core generator operates through three interconnected modules:

\begin{figure}[!t]
    \centering
    \includegraphics[width=0.94\linewidth]{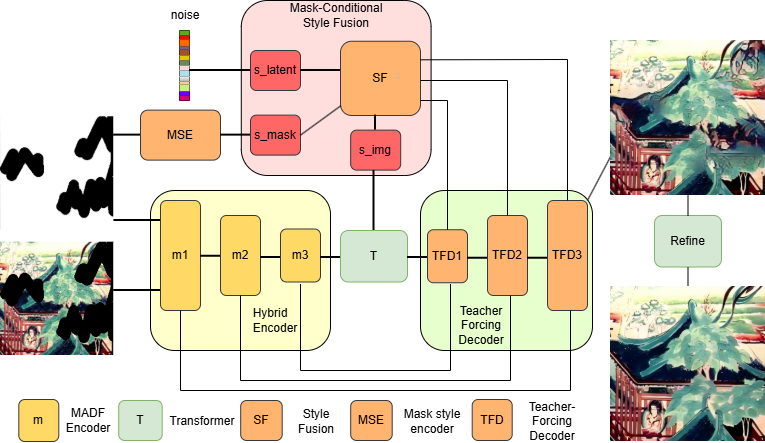} 
    \caption{Overview of the proposed Hybrid Mask-Aware Transformer (HMAT). The core architecture integrates Hybrid Encoder (MADF + Transformer) for robust feature extraction, Mask-Conditional Style Fusion (SF) to dynamically guide synthesis, and a Teacher-Forcing Decoder (TFD) with hard-gated skip fusion, whose empirical effect is analyzed in the ablation study. The resulting coarse structural completion is composited with the observed pixels and then processed by a second-stage synthesis network to produce the final restored image.}
    \label{fig:architecture}
\end{figure}

\begin{itemize}
    \item \textbf{Hybrid Encoder ($E$):} Extracts multi-scale mask-aware representations from the degraded input. By dynamically adapting to the damage mask, it captures fine-grained boundary information without introducing artifacts, formulated as $F_{enc} = E(x_{obs}, m)$.
    \item \textbf{Mask-Conditional Style Fusion ($S$) \& Transformer ($T$):} The Style Fusion module fuses semantic, latent, and mask-geometry cues into a multi-modal condition vector $s = S(m, z)$. This vector is injected into the decoder through FiLM-style affine modulation, while the Transformer leverages long-range contextual attention to hallucinate missing global structures: $F_{global} = T(F_{enc}, s)$.
    \item \textbf{Teacher-Forcing Decoder ($D$):} Projects the enriched features back into the image space. It uses a hard-gated skip fusion mechanism intended to impose observed-region encoder features as boundary conditions during decoding: $\tilde{x} = D(F_{global}, F_{enc}, m)$.
\end{itemize}

The first stage produces a coarse completion $\hat{x}_c$, which is composited with the observed pixels and passed to a second-stage synthesis network to produce the final output $\hat{x}$. Following the coarse-to-fine design of MAT~\cite{li2022matmaskawaretransformerlarge}, both stages are trained jointly,
and each stage is supervised by its own discriminator. The final output is composited with the observed pixels (Eq.~\ref{eq:composite}) to preserve authentic regions exactly.

\subsection{Hybrid Encoder: Spatially Conditioned Linear Operators}

To address feature corruption caused by irregular missing regions, it is necessary to ensure robust feature extraction near the boundaries between valid and damaged areas. However, a standard convolution applies a spatially invariant kernel $W$ uniformly across the image:
\begin{equation}
y(p) = \sum_{q \in N(p)} W(q)x(p+q).
\end{equation}
When $x$ contains large zero-filled holes, the receptive field $N(p)$ near these boundaries inevitably mixes authentic pixel values with artificial zeros~\cite{suvorov2022resolution}. This leads to biased feature responses and boundary artifacts, which are particularly detrimental in mural restoration, where crack edges and damaged contours often carry important structural information.

Instead of using a fixed convolution kernel, we employ MADF~\cite{Zhu_2021} in the encoder to resolve the issue. MADF generates a location-dependent kernel $W_p$ conditioned on the local mask structure. Specifically, for each spatial location $p$, the kernel is dynamically predicted as
% \begin{equation}
% W_p = \Phi(\text{mask neighborhood around } p),
% \end{equation}
% and the convolution becomes
% \begin{equation}
% y(p) = \sum_{q \in N(p)} W_p(q)x(p+q).
% \end{equation}
\begin{equation}
    W_p = \Phi(m_{\mathcal{N}(p)}),
\end{equation}
where $m_{\mathcal{N}(p)}$ denotes the mask neighborhood around $p$. The
corresponding convolution becomes
\begin{equation}
    y(p) = \sum_{q \in \mathcal{N}(p)} W_p(q)x(p+q).
\end{equation}

This formulation can be interpreted as a spatially varying linear operator $A(m)$ applied to the input, i.e., $y = A(m)x$, where the operator adapts to the geometry of the missing regions. As a result, the encoder is able to reduce feature contamination near irregular boundaries and preserve high-frequency details.

In addition, we propagate a soft validity embedding through the encoder layers, updating it as $m^{(l+1)} = \psi(m^l)$. This mechanism allows the network to maintain awareness of valid regions across multiple scales, further improving robustness to complex degradation patterns.

\subsection{Transformer: Masked Contextual Aggregation}
After the MADF encoder, the extracted feature map is flattened into a sequence of $N$ tokens, $X \in \mathbb{R}^{N \times C}$. We also derive a binary token-validity indicator $v \in \{0,1\}^N$ from the propagated mask, where $v_j=1$ signifies that token $j$ contains valid observed content.

To hallucinate large missing structures, these tokens are processed through the Transformer bottleneck of HMAT. Within each local window~\cite{liu2021swin}, standard self-attention computes similarities $s_{ij} = (q_i \cdot k_j) / \sqrt{d}$, where $q_i$ and $k_j$ represent the query and key vectors, and $d$ is their channel dimension. However, to prevent invalid tokens (those falling entirely inside a peeling hole) from acting as sources of corrupted information, we apply a validity-aware mask bias:
\begin{equation}
    s'_{ij} = 
    \begin{cases} 
      s_{ij}, & \text{if } v_j = 1 \\
      -\infty, & \text{otherwise} 
    \end{cases}
\end{equation}
The attention weights are then computed as $\alpha_{ij} = \text{softmax}_j(s'_{ij})$. This masking makes each updated token a weighted average formed from valid source tokens. Missing tokens are updated by borrowing context from observed, authentic mural regions, enabling long-range semantic reconstruction without being corrupted by the artificial zeros inside the damaged areas. As tokens are updated, the validity mask undergoes morphological dilation in token space ($v_i^{(l+1)} = \max_{j} v_j^{(l)}$), gradually expanding the known context deeper into the unknown blocks.

\subsection{Mask-Conditional Style Fusion}
Missing regions vary considerably in morphology, with thin cracks requiring continuous texture synthesis, whereas large peeling areas demand substantial structural reconstruction. To address this variability, HMAT employs a conditional generative prior that is dynamically controlled by a fused style code~\cite{zhao2021large}. To be more specific, the Mask-Conditional Style Fusion module extracts three distinct representations: 
\begin{itemize}
    \item \textbf{Semantic Style ($s_{img}$):} Computed via global average pooling of the Transformer features, summarizing the global content and color palette.
    \item \textbf{Latent Style ($s_{latent}$):} Derived from the noise vector $z$ via a mapping network, providing stochastic multi-modality for ambiguous regions.
    \item \textbf{Mask Geometry Style ($s_{mask}$):} Extracted from a dedicated shape encoder applied to $m$, explicitly encoding the geometric morphology of the degradation.
\end{itemize}
These representations are concatenated and passed through a learnable mapping network to form the final style vector:
\begin{equation}
    s=\phi([s_{img},s_{latent},s_{mask}]).
\end{equation}
The style vector supplies complementary semantic, stochastic, and geometric conditioning for adapting the generator to different degradation patterns. It modulates the first-stage decoder via StyleGAN2-style weight modulation~\cite{karras2020stylegan2}. For each decoder convolution, an affine
layer maps $s$ to per-input-channel scales $s_i$, which modulate and demodulate the convolution weights:
\begin{equation}
    w'_{oik}=s_i\,w_{oik},\qquad
    w''_{oik}=\frac{w'_{oik}}{\sqrt{\sum_{i',k'}(w'_{oi'k'})^2+\epsilon}}.
\end{equation}
where $o$, $i$, and $k$ denote the output channel, input channel, and spatial kernel index, respectively. This modulation allows the fused style code to guide the decoder toward texture completion for thin cracks or structural reconstruction for larger peeling regions.

\subsection{Teacher-Forcing Decoder}
Within the first-stage decoder, encoder skip features are fused with decoder features through a hard gate rather than standard addition. This design is intended to make observed-region encoder features act as boundary conditions while the decoder synthesizes missing-region content. Let $D_l$ and $S_l$ denote the decoder and encoder-skip features at scale $l$, and let $m_l$ denote the validity mask resized to the same scale:
\begin{equation}
    F_l = D_l \odot (1 - m_l) + S_l \odot m_l .
\end{equation}
At valid locations, the fused feature follows the encoder skip feature, i.e., $F_l(p)=S_l(p)$ when $m_l(p)=1$. In this case, no gradient reaches the decoder feature at that location, $\partial F_l(p)/\partial D_l(p)=0$. The decoder therefore treats $\bar{\Omega}$ as a feature-space boundary condition and focuses synthesis on the missing region $\Omega$. We analyze the empirical contribution of this gate in the ablation study in Section~\ref{sec:ablation}.

\subsection{Training Objective}
\label{sec:objective}
HMAT is trained jointly end-to-end (both generator stages, discriminators, and all losses). Recall that $\hat{x}$ is the final restored image and $\hat{x}_c$ the coarse first-stage output, both composited (Eq.~\ref{eq:composite}) so observed pixels equal $x_{obs}$ exactly.

\noindent\textbf{Reconstruction.} Because compositing makes valid pixels exact,
we apply a hole-normalized $\ell_1$ term that penalizes only the missing region
$\Omega$, on both the coarse and final outputs:
\begin{equation}
    \mathcal{L}_{rec} =
    \frac{\lVert (\hat{x}-x)\odot(1-m)\rVert_1}{C\,\lvert\Omega\rvert+\epsilon}
    +
    \frac{\lVert (\hat{x}_{c}-x)\odot(1-m)\rVert_1}{C\,\lvert\Omega\rvert+\epsilon},
\end{equation}
where $C$ is the number of channels and $\lvert\Omega\rvert$ the number of missing pixels.

\noindent\textbf{High-receptive-field perceptual loss.} Following
\cite{suvorov2022resolution}, we match deep features from a frozen
dilated ResNet-50 segmentation network $\Phi$ pretrained on ADE20K, whose large
receptive field rewards globally coherent structure:
\begin{equation}
    \mathcal{L}_{hrf} = \sum_{k}\big\lVert \Phi_k(\hat{x})-\Phi_k(x)\big\rVert_2^2 .
\end{equation}

\noindent\textbf{Adversarial and feature-matching losses.} We use a two-scale
discriminator $D$ (operating on $\hat{x}$ and $\hat{x}_{c}$) with the
non-saturating logistic objective and a discriminator feature-matching term that
aligns intermediate activations $D^{(l)}$ of real and restored images:
\begin{equation}
    \mathcal{L}_{adv}^{G} =
    \mathbb{E}\big[\,\mathrm{sp}(-D(\hat{x})) + \mathrm{sp}(-D(\hat{x}_{c}))\,\big],
    \quad
    \mathcal{L}_{fm} = \frac{1}{L}\sum_{l=1}^{L}
    \big\lVert D^{(l)}(\hat{x})-D^{(l)}(x)\big\rVert_1,
\end{equation}
where $\mathrm{sp}(\cdot)=\log(1+e^{\cdot})$ is the softplus function and the sum
runs over the $L$ tapped layers of both discriminator scales.

\noindent The total generator loss is
\begin{equation}
    \mathcal{L}_{G} = \mathcal{L}_{adv}^{G}
    + \lambda_{rec}\,\mathcal{L}_{rec}
    + \lambda_{fm}\,\mathcal{L}_{fm}
    + \lambda_{hrf}\,\mathcal{L}_{hrf},
    \label{eq:gloss}
\end{equation}
with $\lambda_{rec}=\lambda_{fm}=\lambda_{hrf}=10$.

\noindent\textbf{Discriminator.} The discriminator is trained with the
non-saturating loss and lazy $R_1$ gradient penalty on both scales:
\begin{equation}
    \mathcal{L}_{D} =
    \mathbb{E}\big[\mathrm{sp}(D(\hat{x}))+\mathrm{sp}(-D(x))\big]
    + \mathbb{E}\big[\mathrm{sp}(D(\hat{x}_{c}))+\mathrm{sp}(-D(x))\big]
    + \frac{\gamma}{2}\,\mathbb{E}\big[\lVert\nabla_{x} D(x)\rVert_2^2\big],
\end{equation}
with $\gamma=2$ applied every 16 discriminator steps. We adopt a two
time-scale update rule (TTUR), using Adam ($\beta_1{=}0,\ \beta_2{=}0.99$) with
generator and discriminator learning rates of $10^{-3}$ and $10^{-4}$,
respectively.

\noindent\textbf{Objective analysis.} The proposed objective is designed to make
the learning signal consistent with the restoration protocol. Since the final
output is composited with the observed pixels, applying reconstruction loss over
the full image would over-emphasize already preserved valid regions and make the
effective supervision depend strongly on mask size. The hole-normalized
reconstruction term instead rescales the pixel loss by the number of missing
pixels, producing a comparable optimization scale across moderate and severe
damage patterns. This is important for mural restoration, where the model must
handle both thin cracks and large peeling regions within the same training
distribution. Meanwhile, the high-receptive-field perceptual loss complements the
local pixel term by encouraging globally coherent line structures and semantic
layout, and the discriminator feature-matching loss stabilizes adversarial
training by aligning intermediate statistics of real and restored murals. These
three terms therefore provide complementary supervision: hole-normalized
reconstruction enforces faithful damaged-region recovery, high-receptive-field
perception encourages structural coherence, and feature matching improves texture
realism without relying solely on the adversarial signal.

\section{Experiments} \label{experiment}

\subsection{Datasets and Metrics}
\label{sec:experiment}
We evaluate HMAT on two datasets with different restoration challenges. First, we use the DHMural dataset~\cite{shao2023}, which contains 13,233 mural images. Following the official protocol, we use 10,584 images for training and 2,649 for testing, all resized to $256 \times 256$. Second, to assess performance on intricate line-drawing styles, we construct a Nine-Colored Deer dataset by cropping a high-resolution gigapixel scan into 4,000 non-overlapping $256 \times 256$ patches, with 3,500 for training and 500 for testing. To simulate realistic weathering, we adopt a Random Brush Stroke algorithm~\cite{yu2019free} to generate irregular free-form masks that mimic peeling paint and cracks. We further divide the test masks into Moderate (20--30\% coverage) and Severe (40--50\% coverage) settings to evaluate robustness under different degradation levels. Masks whose coverage falls outside the target range are discarded and regenerated until the predefined Moderate and Severe test sets are obtained.

For reproducible evaluation, the random masks are generated once and then saved as fixed test mask sets. The same saved Moderate and Severe mask sets are used for HMAT and all baseline methods. Each mask setting is evaluated on the full test split of each dataset, resulting in 2,649 test samples for DHMural and 500 test samples for Nine-Colored Deer under each mask coverage range. This protocol ensures that performance differences are caused by the restoration models rather than by variation in randomly sampled masks.

For quantitative evaluation, we use PSNR, SSIM, FID, and $\ell_1$ Error to assess both fidelity and realism. PSNR and SSIM measure signal fidelity and structural consistency, which are important for preserving original line work. FID evaluates the perceptual realism of the generated textures with respect to the real data distribution. In our evaluation, FID is computed on the full composited restoration outputs. $\ell_1$ Error measures pixel-wise absolute deviation, reflecting the accuracy of restored color values against the ground truth. Following standard inpainting evaluation, full-image metrics are computed on the composited restoration output, where observed pixels are copied from the input and only missing regions are synthesized. Since full-image scores can be dominated by unchanged valid pixels, we additionally report hole-region and boundary-band metrics in Table~\ref{tab:ablation} to isolate synthesized-region fidelity and local blending quality.

\subsection{Implementation Details}
We adopt a compact architecture to balance computational efficiency and dense feature modeling for mural restoration. The MADF encoder uses a channel schedule of $[64,128,180]$, projecting features to a 180-dimensional bottleneck. This design provides richer semantic capacity than 128 channels while avoiding the memory cost of wider settings such as 256. The encoded features are then processed by a Transformer bottleneck with 5 blocks and 8 attention heads, using the same hidden dimension of 180 to model long-range structure.

The model is trained at $256 \times 256$ resolution on 2 RTX3090 GPUs with a batch size of 4. We use Adam with a two time-scale update rule (TTUR), setting the generator and discriminator learning rates to $10^{-3}$ and $10^{-4}$, respectively. Geometric augmentation is disabled to preserve the authentic orientation of historical brushstrokes and line patterns. For a fair comparison, all learning-based baselines, including MADF, MAT, LaMa-Fourier, and EdgeConnect, are reproduced using their official codebases and retrained on the same training splits as HMAT. All methods are trained and evaluated at $256 \times 256$ resolution using the same Moderate and Severe mask sets, the same train/test splits, and the same output compositing protocol before metric computation. For each method, the checkpoint used for testing is selected according to validation performance on the corresponding training split. Unless otherwise specified by the official implementation, we keep the original optimizer settings of each baseline and only adapt the data loader, image resolution, and mask protocol to our mural restoration setting. All reported results are computed using the same saved test masks to ensure deterministic evaluation.

Following Section~\ref{sec:objective}, we supervise both the coarse output $\hat{x}_c$ and the final output $\hat{x}$. The generator is optimized with the hole-normalized reconstruction loss, high-receptive-field perceptual loss, non-saturating adversarial loss, and discriminator feature-matching loss defined in Eq.~\ref{eq:gloss}. The discriminator is trained with the non-saturating logistic loss and lazy $R_1$ regularization, with $\gamma=2$ applied every 16 discriminator steps.

\subsection{Optimization of Style Fusion Parameters}

The Mask-Conditional Style Fusion module controls the generative prior by combining three complementary representations: Semantic Style ($s_{img}$), Latent Style ($s_{latent}$), and Mask Geometry Style ($s_{mask}$). The baseline dimensionality is selected from a small validation sweep rather than fixed arbitrarily. We first allocate more capacity $s_{img}$ because the mural color palette and global semantic layout are shared across many patches, keep sufficient capacity in $s_{latent}$ to preserve stochastic texture variation, and use a compact $s_{mask}$ branch because the binary mask provides low-entropy geometric information. To evaluate the effect of this capacity distribution, we compare three-dimensional configurations on the Nine-Colored Deer dataset:

\begin{itemize}
    \item \textbf{Equal Capacity:} A flat dimensional distribution where $s_{img} = 180$, $s_{latent} = 180$, and $s_{mask} = 180$.
    \item \textbf{Heavy Semantic Bias:} A highly constrained conditional setup where semantic features dominate ($s_{img} = 360$), while the latent and mask capacities are bottlenecked ($s_{latent} = 64$, $s_{mask} = 16$).
    \item \textbf{Baseline Capacity:} Our baseline distribution ($s_{img} = 360$, $s_{latent} = 180$, $s_{mask} = 64$).
\end{itemize}

\begin{table}[H]
\centering
\caption{Quantitative ablation on the dimensionality of the Mask-Conditional Style Fusion module.}
\label{tab:dimension_ablation}
\small
\setlength{\tabcolsep}{5pt}
\begin{tabular}{lcccc}
\hline
\textbf{Configuration} & \textbf{PSNR $\uparrow$} & \textbf{SSIM $\uparrow$} & \textbf{FID $\downarrow$} & \textbf{$\ell_1$ $\downarrow$} \\ \hline
Equal Capacity & 27.22 & 0.907 & 11.16 & 0.014 \\
Heavy Semantic Bias & 26.58 & 0.896 & 13.79 & 0.016 \\
\textbf{Baseline (Ours)} & \textbf{28.47} & \textbf{0.919} & \textbf{10.90} & \textbf{0.012} \\ \hline
\end{tabular}
\end{table}

\begin{figure}[H]
    \centering
    \includegraphics[width=1\linewidth]{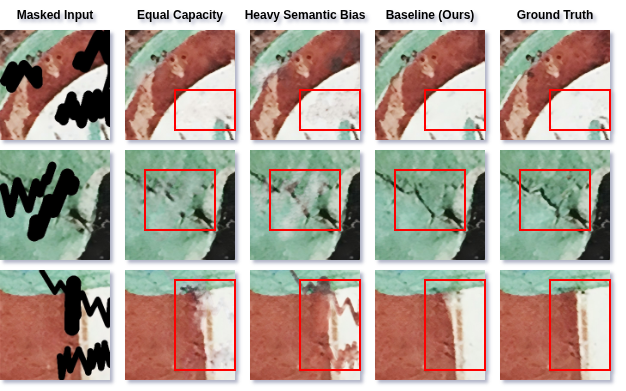} 
    \caption{Qualitative comparison of style dimensionality configurations on the Nine-Colored Deer dataset. To evaluate capacity distribution, we compare our \textit{Baseline} ($s_{img}=360, s_{latent}=180, s_{mask}=64$) against \textit{Equal Capacity} ($s_{img}=180, s_{latent}=180, s_{mask}=180$) and \textit{Heavy Semantic Bias} ($s_{img}=360, s_{latent}=64, s_{mask}=16$).}
    \label{fig:style_ablation}
\end{figure}

As shown in Table~\ref{tab:dimension_ablation} and Figure~\ref{fig:style_ablation}, deviating from the baseline capacity distribution degrades performance. The \textit{Equal Capacity} setting reduces PSNR to 27.22 and increases $\ell_1$ error to 0.014, suggesting that over-emphasizing mask geometry introduces color shifts and boundary artifacts. The \textit{Heavy Semantic Bias} setting performs worst overall, especially in FID (13.79), because bottlenecking the latent and mask branches limits the capacity needed for rich, high-frequency texture synthesis. These results indicate that the baseline distribution is the best-performing configuration among the tested candidates and provides a practical balance between global semantic guidance, stochastic texture variation, and mask-geometry conditioning.

\FloatBarrier

\subsection{Comparison with State-of-the-Art Methods}

We compare HMAT with representative convolutional, transformer-based, large-mask, and edge-guided inpainting baselines, including MADF~\cite{Zhu_2021}, MAT~\cite{li2022matmaskawaretransformerlarge}, LaMa-Fourier~\cite{suvorov2022resolution}, and EdgeConnect~\cite{nazeri19}. To ensure a controlled comparison, all baselines are evaluated under the same input masks, test splits, image resolution, and output compositing protocol as HMAT. Although ADF~\cite{shao2023} is closely related to mural restoration, it addresses blind restoration of globally degraded murals rather than mask-conditioned completion with specified missing regions. We therefore discuss ADF as domain-related work in Section~\ref{relate}, while focusing the controlled benchmark in Table~\ref{tab:sota} on methods that solve the same mask-guided inpainting task. The table reports composited full-image metrics, which measure the quality of the final restored outputs, and the region-specific ablation in Table~\ref{tab:ablation} further isolates synthesis quality within damaged regions and along restoration boundaries.

\begin{table}[!htbp]
\centering
\footnotesize
\caption{Quantitative comparison with state-of-the-art methods on the DHMural and Nine-Colored Deer datasets. Metrics are computed on composited full images.}
\setlength{\tabcolsep}{4pt}
\label{tab:sota}

\textbf{DHMural Dataset}

\vspace{2pt}
\resizebox{\textwidth}{!}{
\begin{tabular}{l|cccc|cccc}
\hline
 & \multicolumn{4}{c|}{Moderate Mask (20\%--30\%)} 
 & \multicolumn{4}{c}{Severe Mask (40\%--50\%)} \\ \cline{2-9}
Method 
& \textbf{PSNR} $\uparrow$ & \textbf{SSIM} $\uparrow$ & \textbf{FID} $\downarrow$ & \textbf{$\ell_1$} $\downarrow$ 
& \textbf{PSNR} $\uparrow$ & \textbf{SSIM} $\uparrow$ & \textbf{FID} $\downarrow$ & \textbf{$\ell_1$} $\downarrow$ \\ 
\hline
MADF & 24.03 & 0.887 & 18.06 & 0.022 & 19.72 & 0.785 & 31.20 & 0.049 \\
MAT & 22.53 & 0.862 & 33.30 & 0.027 & 18.95 & 0.742 & 42.50 & 0.054 \\
LaMa-Fourier & \textbf{25.30} & 0.898 & 17.77 & \textbf{0.019} & 21.35 & 0.812 & 28.40 & 0.042 \\
EdgeConnect & 23.75 & 0.881 & 32.54 & 0.024 & 19.30 & 0.766 & 44.80 & 0.053 \\ 
\hline
\textbf{HMAT (Ours)} & \textbf{25.30} & \textbf{0.899} & \textbf{12.54} & \textbf{0.019} & \textbf{21.85} & \textbf{0.825} & \textbf{22.30} & \textbf{0.039} \\ 
\hline
\end{tabular}
}

\vspace{6pt}
\textbf{Nine-Colored Deer Dataset}

\vspace{2pt}
\resizebox{\textwidth}{!}{
\begin{tabular}{l|cccc|cccc}
\hline
 & \multicolumn{4}{c|}{Moderate Mask (20\%--30\%)} 
 & \multicolumn{4}{c}{Severe Mask (40\%--50\%)} \\ \cline{2-9}
Method 
& \textbf{PSNR} $\uparrow$ & \textbf{SSIM} $\uparrow$ & \textbf{FID} $\downarrow$ & \textbf{$\ell_1$} $\downarrow$ 
& \textbf{PSNR} $\uparrow$ & \textbf{SSIM} $\uparrow$ & \textbf{FID} $\downarrow$ & \textbf{$\ell_1$} $\downarrow$ \\ 
\hline
MADF & 27.05 & 0.898 & 15.80 & 0.015 & 24.20 & 0.815 & 24.60 & 0.028 \\
MAT & 26.53 & 0.898 & 17.20 & 0.015 & 24.02 & 0.813 & 26.10 & 0.029 \\
LaMa-Fourier & 28.10 & 0.914 & 13.60 & 0.013 & 25.55 & 0.846 & 21.20 & 0.024 \\
EdgeConnect & 26.90 & 0.901 & 22.40 & 0.015 & 24.10 & 0.818 & 33.50 & 0.029 \\ 
\hline
\textbf{HMAT (Ours)} & \textbf{28.47} & \textbf{0.919} & \textbf{10.90} & \textbf{0.012} & \textbf{26.05} & \textbf{0.857} & \textbf{17.40} & \textbf{0.022} \\ 
\hline
\end{tabular}
}
\end{table}

On the DHMural dataset, HMAT obtains the best or tied-best pixel-level fidelity and substantially improves perceptual realism. Under moderate masks, our method reaches 25.30 PSNR and 0.899 SSIM, matching LaMa-Fourier in PSNR and $\ell_1$ error while reducing FID from 17.77 to 12.54. Under severe masks, HMAT achieves the best results across all four metrics, with 21.85 PSNR, 0.825 SSIM, 22.30 FID, and 0.039 $\ell_1$ error. These results show that the proposed hybrid architecture improves both reconstruction fidelity and perceptual quality, especially when larger missing regions require broader contextual reasoning.

On the Nine-Colored Deer dataset, the advantage of HMAT is more pronounced. Since all samples are cropped from the same mural, the dataset contains a consistent artistic style and many long, continuous line structures, making it suitable for evaluating structural completion. Under moderate masks, HMAT achieves 28.47 PSNR, 0.919 SSIM, 10.90 FID, and 0.012 $\ell_1$ error, outperforming all baselines. Under severe masks, it obtains the best results across all metrics, reaching 26.05 PSNR, 0.857 SSIM, 17.40 FID, and 0.022 $\ell_1$ error. Compared with convolution-oriented methods such as MADF and LaMa-Fourier, HMAT better restores long-range geometric continuity; compared with the pure Transformer baseline MAT, the MADF encoder and mask-conditional synthesis path help preserve sharper local details and reduce texture degradation.

\begin{figure}[!htbp]
    \centering
    \includegraphics[width=1\linewidth]{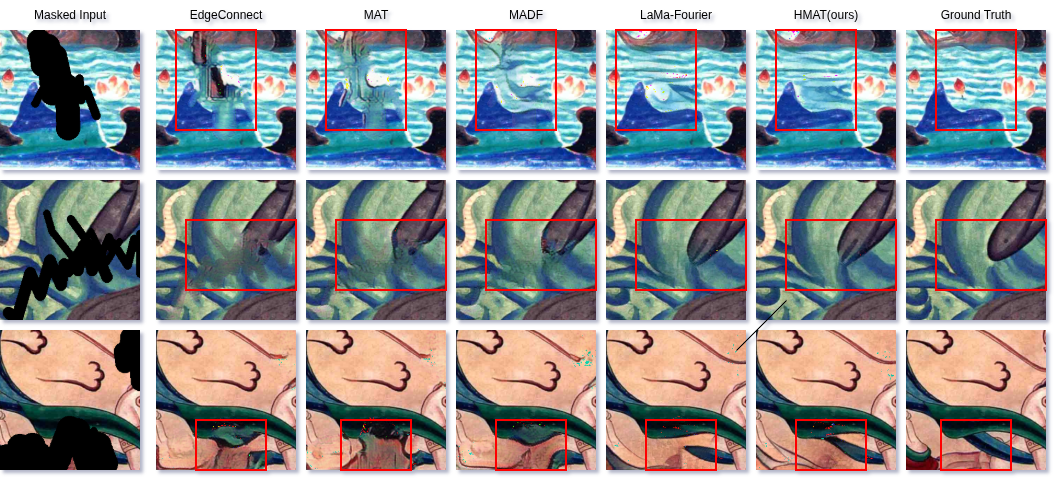}
    \caption{Qualitative comparison with state-of-the-art methods on the DHMural dataset. Red boxes indicate representative regions where competing methods exhibit broken contours, boundary artifacts, or texture degradation.}
    \label{fig:comparison}
\end{figure}

Qualitative comparisons in Figure~\ref{fig:comparison} further support the quantitative results. EdgeConnect and MAT tend to produce unstable textures or broken structures, while MADF and LaMa-Fourier better preserve local patterns but may struggle when large missing regions interrupt long mural contours. In contrast, HMAT produces more coherent lines and visually plausible textures, demonstrating a stronger balance between local texture fidelity and global structural completion.

\FloatBarrier

\subsection{Ablation Study}
\label{sec:ablation}
To evaluate the contribution of each component, we conduct an ablation study on the DHMural test set. Instead of reporting raw scores, Table~\ref{tab:ablation} reports metric changes relative to the full HMAT model. We evaluate both the boundary band around damaged regions and the masked hole region, since boundary metrics reflect local blending quality while hole metrics measure the fidelity of synthesized content.

\begin{table}[!htbp]
\centering
\footnotesize
\setlength{\tabcolsep}{4pt}
\caption{Ablation study on the DHMural test set. We report metric changes relative to the full HMAT model. Positive values indicate improvement, while negative values indicate degradation.}
\label{tab:ablation}
\begin{tabular}{lcccc}
\hline
\textbf{Component} 
& \makecell{\textbf{$\Delta$ Band}\\\textbf{PSNR}} 
& \makecell{\textbf{$\Delta$ Band}\\\textbf{SSIM}} 
& \makecell{\textbf{$\Delta$ Hole}\\\textbf{PSNR}} 
& \makecell{\textbf{$\Delta$ Hole}\\\textbf{SSIM}} \\
\hline
Full HMAT & 0.0000 & 0.0000 & 0.0000 & 0.0000 \\
w/o MADF & -0.3991 & -0.0039 & -0.1996 & -0.0085 \\
w/o Teacher Forcing & +0.0016 & +0.0002 & +0.0009 & +0.0002 \\
w/o Mask-Guided Style & -0.2187 & -0.0024 & -0.1249 & -0.0062 \\
w/o Proposed Objective & -1.1467 & -0.0161 & -1.1108 & -0.0450 \\
\hline
\end{tabular}
\end{table}

As shown in Table~\ref{tab:ablation}, the training objective is the most influential factor. In the ``w/o Proposed Objective'' variant, we replace the fidelity-oriented training objective with a conventional inpainting objective consisting of adversarial loss, VGG-based perceptual loss, and unnormalized $\ell_1$ reconstruction loss. This variant causes the largest degradation, reducing band PSNR by 1.1467 and hole PSNR by 1.1108. The result confirms that the hole-normalized reconstruction loss, high-receptive-field perceptual loss, and discriminator feature-matching loss are critical for faithful mural restoration.

The relative magnitudes of the ablation results also reveal how different components interact. The proposed objective has the largest effect because it directly changes the supervision applied to the synthesized region: hole-normalized reconstruction stabilizes learning across different mask sizes, high-receptive-field perceptual supervision encourages coherent long-range structure, and feature matching improves texture realism. In contrast, the architectural components mainly affect how information is represented and conditioned before synthesis. MADF provides the second-largest contribution by reducing feature contamination near irregular mask boundaries, which benefits both boundary-band blending and hole-region reconstruction. The mask-guided style branch produces a smaller but consistent drop when removed, suggesting that mask geometry is most useful as an adaptive conditioning signal for different degradation morphologies rather than as the sole source of restoration quality.

In contrast, removing the teacher-forcing decoder does not degrade performance. This result suggests that hard-gated skip fusion is not a key contributor under the current training setting. Since output-level compositing already preserves valid pixels exactly and the hole-normalized objective directly focuses supervision on missing regions, explicit teacher forcing provides limited additional benefit. Therefore, the main empirical gains of HMAT come from the improved training objective, MADF-based mask-aware encoding, and mask-guided style conditioning.

\section{Limitations and Future Work}

Although HMAT improves both fidelity and perceptual realism under moderate and severe mask settings, several limitations remain. First, the framework is trained and evaluated at $256 \times 256$ resolution for controlled comparison with existing inpainting baselines, which may not fully capture ultra-high-resolution mural details such as subtle pigment transitions, brushstroke granularity, and fine crack structures. Extending HMAT through tiled inference, multi-scale generation, or more efficient high-resolution restoration strategies~\cite{zhang2025ultra} is a natural direction. Second, the Nine-Colored Deer dataset is constructed from a single mural source, providing a controlled setting for evaluating long-range line continuity but limited coverage of mural styles, historical periods, and material conditions. Building larger and professionally curated benchmarks with museums or conservation institutes would improve evaluation diversity. Third, HMAT is formulated as a mask-conditioned restoration method and assumes that damaged regions are provided as input masks. In practical conservation workflows, such masks may come from manual annotation, hyperspectral analysis, or a separate damage detection model. It does not directly address blind degradation localization, where cracks, peeling, mold, or pigment loss must first be detected. Integrating automatic damage detection or jointly learning localization and restoration is important for practical deployment. Fourth, while output-level compositing guarantees exact preservation of observed pixels, highly ambiguous missing regions may still admit multiple plausible restorations. Expert feedback or uncertainty estimation~\cite{giakoumoglou2025sagi} could make the restoration process more reliable for real-world conservation practice.

\section{Conclusion} \label{conclusion}
We proposed HMAT, a hybrid mask-aware framework for ancient mural restoration that combines MADF-based local texture modeling, Transformer-based structural reasoning, mask-conditional style fusion, and a fidelity-oriented training objective. Experiments on the DHMural and Nine-Colored Deer datasets show that HMAT improves restoration fidelity and perceptual realism under moderate and severe degradation. Ablation studies further identify the proposed objective, MADF-based encoding, and mask-guided style conditioning as the main contributors, while hard-gated teacher-forcing skip fusion provides limited additional benefit under the current setting. By recovering continuous geometric line structures across damaged regions while preserving fine historical textures, HMAT provides an effective solution for digital cultural heritage restoration.

\bibliographystyle{splncs04}
\bibliography{references}
\end{document}